\documentclass[twoside,11pt]{article}

\usepackage{blindtext}

\usepackage[abbrvbib, preprint]{dmlr2e}

\usepackage{lastpage}
\dmlrheading{23}{2024}{1-\pageref{LastPage}}{9/23; Revised 10/23}{11/23}{21-0000}{Nikolaos Myrtakis, Ioannis Tsamardinos and Vassilis Christophides} % Insert the dates and author names. This is not relevant if it is yet being reviewed, i.e., \documentclass[twoside,11pt, preprint]{article}
\ShortHeadings{A Comparative Analysis of Influence Signals for Data Debugging}{A Comparative Analysis of Influence Signals for Data Debugging}
\firstpageno{1}

\usepackage[utf8]{inputenc}
\usepackage{soul}
\usepackage{xcolor}

\usepackage{amsmath}
\usepackage{hyperref}
\usepackage{mathrsfs}
\usepackage{graphicx}
\usepackage{url}

\usepackage{mathtools}

\usepackage{multirow}
\usepackage{diagbox}

\usepackage{pifont}

\usepackage{adjustbox}

\usepackage{subcaption}

\usepackage{dsfont}

\usepackage{txfonts}

\usepackage[ruled,linesnumbered]{algorithm2e}

\SetKwComment{Comment}{/* }{ */}
\begin{document}

\title{A Comparative Analysis of Influence Signals for Data Debugging \thanks{Accepted and presented at the \href{https://icml.cc/virtual/2024/36390}{Data-centric Machine Learning Research (DMLR)} Workshop at ICML 2024.}}

\author{\name Nikolaos Myrtakis \email myrtakis@csd.uoc.gr \\
       \addr Department of Computer Science \& ETIS Lab\\
       University of Crete, Greece \& ENSEA, France\\
       \AND
       \name Ioannis Tsamardinos \email tsamard.it@gmail.com \\
       \addr Department of Computer Science\\
       University of Crete, Greece\\
       \AND
       \name Vassilis Christophides \email vassilis.christophides@ensea.fr \\
       \addr ETIS Lab\\
       ENSEA, France\\
       }

\editor{My editor}

\maketitle
\vspace{-0.8cm}
\begin{abstract}%   <- trailing '%' for backward compatibility of .sty file
Improving the quality of training samples is crucial for improving the reliability and performance of ML models. In this paper, we conduct a comparative evaluation of influence-based signals for debugging training data. These signals can potentially identify both mislabeled and anomalous samples from a potentially noisy training set as we build the models and hence alleviate the need for dedicated glitch detectors. Although several influence-based signals (e.g., Self-Influence, Average Absolute Influence, Marginal Influence, GD-class) have been recently proposed in the literature, there are no experimental studies for assessing their power in detecting different glitch types (e.g., mislabeled and anomalous samples) under a common influence estimator (e.g., TraceIn) for different data modalities (image and tabular), and deep learning models (trained from scratch or foundation). Through extensive experiments, we show that signals like Self-Influence effectively detect mislabeled samples but none of the existing signals can detect anomalies. Existing signals do not take into account the training dynamics, i.e., how the samples' influence on the model changes during training, while some signals fall into influence cancellation effects i.e., influence score is zero due to unsigned scores accumulation, resulting in misleading influence attribution.
\end{abstract}

\begin{keywords}
  Data Debugging, Mislabeled and Anomalous Samples, Influence Functions, Signals
\end{keywords}

\section{Introduction}

It is widely recognized that the success of Machine Learning (ML) systems heavily depends on the quality of data used to train them. Even the most advanced learning algorithms trained on dirty datasets may lead to models that make inaccurate predictions and do not generalize well in real-world settings \citep{Gupta21,Whang21,Chai22}. 

In particular, we are focusing on debugging data glitches that break the following assumptions made by traditional ML algorithms, such as: (a) the observed labels of training samples are correct \citep{song2022learning}; (b) the feature values of training samples are consistent and occupy different regions of the input data space per class \citep{das2018handling}. Unfortunately, these assumptions are rarely met in real-world settings jeopardizing the performance of ML models with nefarious consequences for business applications and people’s lives\footnote{\url{https://www.forbes.com/sites/gilpress/2021/06/16/andrew-ng-launches-a-campaign-for-data-centric-ai/}}. As a matter of fact, the ratio of corrupted labels in real-world datasets is estimated to be between 8.0\% and 20\% \citep{li2017webvision,song2019selfie}.

In this paper, we perform the first comparative evaluation of different signals that have been proposed for debugging data using Influence Functions (IFs) \citep{koh2017understanding}. IFs assess the influence of each training sample on a classification model trained with gradient descent or similar variants. Although several IFs have been recently proposed \citep{hammoudeh2022training}, the signals that are built upon them for identifying the most detrimental samples from a potentially dirty training set have not been thoroughly evaluated for different data glitch types\footnote{We use the term data glitch to denote any systematic change introduced in the data by causes external to the process that generates them and is different from the normal level of random noise present in most data sets \citep{Dasu00}.} and data modalities. Detecting glitches via influence signals does not explicitly solve an anomaly or a mislabeled detection problem. Instead, the model is trained directly to the potentially ``glitched'' dataset, and the influence signals identify the data imperfections during training. Note that these signals can be applied to any parametric ML model trained with gradient descent or similar variants.  

Despite current benchmarking efforts in data debugging (e.g., \citep{mazumder2024dataperf}), there is no previous empirical study addressing how influence-based signals can effectively detect both mislabeled (uniform class vs class-dependent noise) and anomalous (far vs near clusters) samples under a common influence estimator (e.g., TraceIn) for tabular and image data modalities modeled using DL models (e.g., FT-Transformer, ResNet, and MLP), and vision foundation models (e.g., ResNet20, ConvNeXt, and ViT-16B). In a nutshell, the main contributions of our work are:
\begin{itemize}
    % \item To identify the mislabelled, anomalous, and mixed types of glitches we introduce three novel signals, called CNCI, PCID, and CFRank, respectively. Conceptually, we categorize the signals into a new family of signals, that we call {\em counterfactual} ones, examining the change of a sample's influence if its label had been different. 
%    \item We introduce the first taxonomy of the influence-based signals as defined in prior works and a unified notation that can facilitate the introduction of novel influence-based signals.
    \item We conduct the first comparative evaluation of the four influence-based signals namely Self Influence, 
% \citep{koh2017understanding,pruthi2020estimating,kong2021understanding},
    Average Absolute Influence, % \citep{hara2019data}, 
    Marginal Influence, 
    % \citep{chen2021hydra,kong2021resolving}, 
    and GD-class 
    % \citep{nguyenCBIF}) 
    w.r.t. their detection performance of data glitches under the same influence estimator for a fair comparison.
    \item We assess the power of signals to detect (i) mislabeled samples and for the first time (ii) Near-Clustered Anomalies i.e., samples of the minority class in tabular datasets, (iii) Far-Clustered Anomalies, and (iv) outliers, created via data corruptions in the training set of image datasets. 
    \item We apply the four signals to powerful and popular deep learning models either trained from scratch (for tabular data) or pretrained foundation models (for image data).
    % namely ResNet-20 \citep{he2016deep}, ConvNeXt \citep{liu2022convnet} and ViT-16B \citep{DosovitskiyB0WZ21} and benchmark image datasets, namely MNIST, Fashion-MNIST and CIFAR-10 ,\mycomment{to be completed with tabular}.  
    \item We highlight the strengths and weaknesses of each signal showing that current signals such as Self-Influence can effectively detect uniform class noise with more than 10\% noise ratio but none of the existing signals can detect anomalies.
    \item We discuss methodological and foundational research advances in influence-based detection towards an \emph{Explainable Data Debugging} Framework that could detect, characterize, and repair arbitrary glitch mixtures in training sets using influence signals.
    % \item We show that for low-class noise ratio, CNCI outperforms the state-of-the-art mislabeled detection signal, \emph{Self-Influence}, by a margin of 65\% (F1) on average, as well as the mislabel detector CleanLab\citep{cleanlab} by a margin of 75\% (F1) on average. For anomalies, PCID outperforms up to a factor of 6 the existing influence signals, as well as 3 anomaly detectors (Isolation Forest \citep{liu2008isolation}, Deep-SVDD \citep{ruff18a}, DIF\citep{xu2023deep}) in two out of the three datasets while being the most robust.
\end{itemize}

% Our study provides three major insights: (i) the influence footprint of mislabeled samples to the model is similar to hard samples; however the former affect negatively the samples of their true unknown class, (ii) anomalies proved to be easy-to-learn samples having extremely low positive influence on each foundation model and (iii) mislabeled and anomalies have fundamentally different influence signatures during model training that leads to their detection and characterization.

The remainder of the paper is organized as follows. Section \ref{sec:data-glitches} describes the different types of data glitches (mislabeled and anomalous samples) studied in our work. In Section \ref{sec:influence-sigs} we survey the four influence-based signals. Section \ref{sec:experimental-setting} details the experimental testbed. Section \ref{sec:experimental-evaluation} reports the results of our experimental evaluation and the lessons learned. Finally, Section \ref{sec:discussion} concludes the findings of our work and highlights interesting research paths on influence-based glitch detection.

%%%%%%%%%%%%%%%%%%%%%%%%%%%%%%%%%%%%%%%%%%%%%%%%
%%%% DATA glitches
%%%%%%%%%%%%%%%%%%%%%%%%%%%%%%%%%%%%%%%%%%%%%%%%
\section{Data glitches}
\label{sec:data-glitches}

In this section, we describe the main types of data glitches that we address in this work, namely, \emph{mislabeled} and \emph{anomalous} samples. Some data glitches may arise due to human or script errors in the data acquisition, transmission, and collection processes, while others are a natural product of the intrinsic nature of the domains of interest. We should stress that we focus on \emph{non-deliberate} data glitches in \emph{training} sets that may incur systematic bias in ML models. Deliberate glitches such as data poisoning and backdoor attacks \citep{tian2022comprehensive} in training sets or test set errors \citep{liu2021towards} are beyond the scope of this work. 

\subsection{Mislabeled samples}
\label{sec:mislabeled}

Let $D_{train}$ be a training dataset $D_{train}=\{(x,\tilde{y})\}_{i=1}^n$, comprised of $n$ pairs of feature vectors $x \in \mathbb{R}^d$ and possibly noisy (observed) labels $\tilde{y} \in \mathbb{Z}$. We denote $C \in \mathbb{Z}^k$ as the set containing the $k$ classes, i.e., the domain of the noisy label vector $\tilde{Y}$. We denote $y$ as the true unknown class of a sample; hence, if $\tilde{y} \neq y$ the sample is considered {\em mislabelled}. These errors can be caused by human errors, annotation tool faults, or data corruption. In our work, we assume that label errors or class noise are generated from a stochastic process that is either independent or dependent w.r.t. the sample features \citep{song2022learning,oyen2022robustness,Frenay14}. We focus in this empirical study on error generation mechanisms that are conditionally independent of the features given the true class $P(\tilde{y} \vert X, y) = P(\tilde{y} \vert y)$. 
%Hence, Feature-Dependent class noise is beyond the scope of this work.

\emph{Uniform} Class Noise. In this type of noise, also known as \emph{symmetric} noise, the true label of a sample $y = i$ is flipped to another label $\tilde{y} = j$ with equal probability. Given a set of $C$ classes, $P(\tilde{y} = j \vert y = i) = \frac{\epsilon}{|C| - 1}, \forall j \neq i \wedge i,j \in C$ and $P(\tilde{y} = i \vert y = i) = 1-\epsilon, \forall i \in C$, where $\epsilon$ denotes the flip probability.

\emph{Class-Dependent} Noise. In this type of noise, also known as \emph{asymmetric} noise, the true label of a sample is more likely to be flipped to a specific class. Given a set of  classes $C$, class-dependent noise is defined as: $\max_{c \in C \setminus \{i,j\}} P(\tilde{y} = c \vert y = i) < P(\tilde{y} = j \vert y = i) < \epsilon$. Note that in asymmetric noise $P(\tilde{y} = i \vert y = i) = 1-\epsilon, \forall i \in C$.

Although robust models are proved to be tolerant to the uniform class noise, class-dependent noise may pose a serious risk to models' performance resulting in up to $\sim 20\%$ drop in test accuracy for 20\%  noise \citep{oyen2022robustness}. This is particularly important for deep neural networks that can overfit a training set even with a high ratio of corrupted labels due to a large number of parameters, resulting in poor generalization performance \citep{zhang2021understanding}. 

\subsection{Anomalous samples}
\label{sec:anomalies-def}
An anomaly is a sample that is irregular from the remainder of the samples in the dataset. Anomalies can occur due to data entry errors, bugs in data wrangling and preprocessing software, sensor faults, etc. Unlike Out-Of-Distribution (OOD) samples \citep{farquhar2022out} observed in test sets, in this work we study the influence of anomalous samples in a \emph{training} set where their feature representation significantly deviates from the rest of the samples. According to \citep{bishop1994novelty}, an anomaly $x$ should satisfy $p(x) < \lambda$, where $\lambda$ is a density threshold and $p$ the probability density function, indicating that anomalies often lie in sparse, low-density areas. In particular, we consider \emph{clustered anomalies}, i.e., samples sharing common characteristics that have not been previously seen during training (aka novelties), and \emph{outliers}, i.e., samples that have been corrupted under a common corruption mechanism. As frequently observed in tabular data\footnote{\url{https://odds.cs.stonybrook.edu/}} novelties form \emph{far-isolated} clusters while outlier samples are \emph{scattered}. Similarly to OOD literature \citep{yang2022openood}, we consider two types of clustered anomalies (CA): (i) Near-CA and (ii) Far-CA. For the tabular datasets considered in our work, Near-CA are samples of the minority class, i.e., rare events while the Far-CA concern samples from a completely different distribution. For an image dataset, Far-CA can be formed as follows: given a dataset $D$ on which a ML model is trained, a subset of samples of another dataset $D'$ from a specific class $c$ of $D'$, such as $S'_c \subseteq D'_c$ where $D'_c = \{z_i = (x_i, y_i) \in D' | y_i = c\}$. The samples of $S'$ are now part of $D$, i.e., $D = D \cup S'_c$. Instances of $S'_c$ can be for example undetected OOD\footnote{\url{https://github.com/Jingkang50/OpenOOD}} samples slipped into the training set. The objective is to detect and inform human analysts about their presence as they come from a different distribution.

Similarly to mislabeled samples, the effect of anomalies has been studied mainly under the lens of ML model performance. Several empirical studies have shown that anomalies do not significantly affect the model performance \citep{Li21,Neutatz22,hara2019data}. However, the influence of anomalies in the formation of the model's decision boundary is left unexplored. In other words, does the reported insubstantial impact on performance also mean low influence?

%%%%%%%%%%%%%%%%%%%%%%%%%%%%%%%%%%%%%%%%%%%%%%%%
%%%% Influence Signals
%%%%%%%%%%%%%%%%%%%%%%%%%%%%%%%%%%%%%%%%%%%%%%%%

\section{Influence-Based Signals}
\label{sec:influence-sigs}

IFs were first introduced for regression tasks  \citep{hampel1974influence,cook1977detection} and they recently served as model diagnostic tools in classification tasks \citep{koh2017understanding,koh2019accuracy}. Apart from explaining the model's predictions, IFs are valuable for debugging data, mainly detecting uniform mislabeled samples during training \citep{koh2017understanding,pruthi2020estimating,kong2021understanding,nguyenCBIF,hara2019data,teso2021interactive} or testing \citep{thimonier2022tracinad}. IFs are efficient approximations of the \emph{Leave-One-Out-Retraining} (LOOR) semantics. LOOR quantifies the change in the loss\footnote{one may examine the change in predictions or parameters} of a sample $z_j = (x_j, y_j)$ if another sample $z_i = (x_i, y_i)$ is excluded from the training set and the model is retrained to obtain the new optimal parameters. 

Clearly, LOOR is infeasible for models with a moderate size of parameters even on small datasets, as it requires $n+1$ retrainings to convergence, where $n$ is the training set size. For this reason, \emph{gradient-based} IFs have been introduced \citep{koh2017understanding,relatif,kong2021resolving,hara2019data,pruthi2020estimating,kong2021understanding,chen2021hydra,yeh2018representer,sui2021representer} to estimate the \emph{leave-one-out} effect without re-training models. Gradient-based IFs could be either \emph{static} or \emph{dynamic} \citep{hammoudeh2022training}. The former estimates the samples' influence using the final model, i.e., at the end of the training phase. The latter provides a more fine-grained view of influence by unrolling the gradients throughout the training process, capturing the training dynamics.

Despite the wide-range applications of static IFs \citep{reinCleaningBench, kong2021resolving, han2020explaining, cohen2020detecting}, the effectiveness of this category has been criticized by several recent works \citep{BasuPF21, schioppa2023theoretical, zhang2022rethinking, bae2022if}. The main concerns are (i) scalability, as the inverse Hessian matrix must be computed, (ii) convexity, as the influence estimation accuracy of static IFs has been proved to decrease in deep neural networks, and (iii) influence fading, as highly influential samples can appear uninfluential at the end of training \citep{schioppa2022scaling}, which is particularly important for data debugging. Thus, in our study, we evaluate the influence-based signals on dynamic IFs and specifically TracIn \citep{pruthi2020estimating}. TracIn estimates how the parameter updates, caused by a sample $z_i$, affect the loss $\mathcal{L}$ of $z_j$ through time, i.e., across epochs: $\mathcal{I}(z_i \rightarrow z_j)  = \sum_{t=1, z_i \in B_t}^T \frac{\eta_t }{|B_t|} \nabla \mathcal{L}(z_j, \theta_{t}) \nabla \mathcal{L}(z_i, \theta_{t})$, where $B_t$, $\theta_t$ and $\eta_t$ are the sample batch that contains $z_i$ the model parameters and the learning rate at the epoch $t$. Note that the influence of $z_i$ can be positive (helping the classification of $z_j$), negative (degrading the classification performance for $z_j$), or zero (having no effect on $z_j$).

An influence signal is an aggregation over the influence scores computed by an IF. We focus on two main categories of influence signals (i) \emph{self} and (ii) \emph{joint} signals. The (i) refers to the influence of a sample on itself while the (ii) is computed between pairs of samples. Joint signals measure the \emph{train-to-validation} influence between two samples $z_i, z_j$, where $z_i \in D_{train}$ and $z_j \in D_{val}$, assuming that the samples in the validation set are clean. This assumption is followed often in current benchmarking efforts \citep{mazumder2024dataperf} and prior works \citep{zhang2018training,teso2021interactive}. We consider two types of joint signals namely \emph{marginal} and \emph{conditional}. The former is calculated irrespectively of the class of the examined samples and the latter by considering the class information. Subsequently, we survey all the proposed signals leveraging influence information for data debugging. Note that these signals are orthogonal w.r.t. the employed influence estimator.

The seminal signal for detecting mislabeled samples is \emph{Self Influence (SI)} \citep{koh2017understanding, pruthi2020estimating,kong2021understanding}. Considering a sample $z_i$, SI is defined as $\mathcal{I}(z_i \rightarrow z_i)$, providing an estimation of the error incurred on $z_i$ if it is removed from the training set. Samples with high SI magnitude are considered ``influential'' and are more prone to have errors in labels \citep{koh2017understanding,pruthi2020estimating} or features \citep{kong2021understanding}. Although previous studies report that SI can also detect anomalies in generative models \citep{kong2021understanding}, our experiments do not confirm such results for classification tasks. 

\emph{Marginal Influence} (MI) \citep{kong2021resolving,chen2021hydra} is a marginal joint influence signal defined as $\mathcal{I}(z_i \rightarrow S) = \sum_{j=1}^m \mathcal{I}(z_i \rightarrow z_j)$, where $z_j \in D_{val}$, capturing the marginal influence of $z_i$ to the validation. Note that MI does not take the influence sign into account, i.e., negative/positive nor class information in the influence aggregation. 

\emph{Average Absolute Influence} (AAI) \citep{hara2019data} is a marginal joint influence signal, defined as $I(z_i \rightarrow S) = \sum_{j=1}^m \frac{1}{m} |\mathcal{I}(z_i \rightarrow z_j)|$, where $z_j \in D_{val}$. The main advantage of AAI over MI is that it does not suffer from influence cancellation during the aggregation, i.e., positive and negative values that lead to zero cumulative influence. However, similar to MI, AAI does not consider class information.

\emph{GD-class} \citep{nguyenCBIF} is a joint conditional signal defined as $\mathcal{I}(z_j \rightarrow S) = \mathit{min}_{y \in Y} \sum_{j=1}^m \mathcal{I}(z_i \rightarrow z_j| y_i = c \wedge y_j  =c)$, where $z_j \in D_{val}$. Note that the GD-class signal considers the class information but not the influence sign. 

All the aforementioned signals assume that the higher the value, the more likely a sample is ``glitched''. None of the existing influence signals consider the influence sign and the class information simultaneously. Finally, the existing signals are often evaluated on detecting uniform class noise. In this work, we consider three additional glitch types namely class-dependent noise, Near-CA, and Far-CA, showing how different glitches challenge the existing signals.

%%%%%%%%%%%%%%%%%%%%%%%%%%%%%%%%%%%%%%%%%%%%%%%%
%%%% Experimental Setting
%%%%%%%%%%%%%%%%%%%%%%%%%%%%%%%%%%%%%%%%%%%%%%%%
\section{Experimental Setting}
\label{sec:experimental-setting}

All experiments run on a 16-core Intel Xeon, 64 GB of main memory, and no GPU. The code is available on \url{https://github.com/myrtakis/influence-signals-benchmark}.

\textbf{Datasets.} We employ three benchmark image datasets, namely MNIST, Fashion-MNIST, and CIFAR-10, which are widely used in the IF literature \citep{pruthi2020estimating,kong2021resolving,teso2021interactive,chen2021hydra,hara2019data,kong2021understanding}. Regarding the tabular data, we elected three popular datasets with varying sample size, dimensionality namely Jannis, Forest Cover and Epsilon \citep{gorishniy2021revisiting}. Their selection is based on the finding of \citep{gorishniy2021revisiting}; most of the DL models perform well in these datasets and as a result, the accuracy of the influence estimates is increased. For each dataset, we draw a 10\% stratified\footnote{Stratification preserves the initial class distribution when splitting or subsampling} subset uniformly at random to speed up influence computations as performed in \citep{teso2021interactive}. Then we split each subset into 80\% for training and 20\% for validation.

\textbf{Contamination of training sets.} For each $D_{tr}$ (tabular and vision datasets), we introduce uniform class noise, by randomly flipping the label of $D_{tr}$ samples with a small probability $\epsilon$ \citep{koh2017understanding,pruthi2020estimating, nguyenCBIF} to a different class uniformly at random. For the class-dependent noise, we select a class $c_i$ uniformly at random that contains the samples $S_i$ and flip the labels of $S_i$ to another class $c_j$ with a small probability $\epsilon$, as described in Section \ref{sec:mislabeled}. To inject anomalies in vision datasets we follow the methodology described in Section \ref{sec:anomalies-def}. Specifically, to create far-isolated clusters we first select a small fraction of samples from a specific class of MNIST denoted as $S_M$, Fashion-MNIST denoted as $S_{F-M}$, and CIFAR-10 denoted as $S_C$. We then use the following combinations to contaminate the training sets of each dataset: $D_M = D_M \cup S_{F-M}$, $D_{F-M} = D_{F-M} \cup S_{M}$, and $D_C = D_C \cup S_{F-M}$. Note that we assign a random class to the anomaly subset that exists in the contaminated dataset. To inject outliers, we used MNIST-C \citep{mu2019mnist} and CIFAR-10-C \citep{hendrycks2018benchmarking} that include corrupted versions of MNIST and CIFAR-10. In particular, we selected a small fraction of samples and corrupted them using \emph{brightness} and \emph{stripe} corruptions in the training sets generated by \citep{mu2019mnist}. Note that both corruption types impact models' generalization performance \citep{mu2019mnist}. Finally, in tabular datasets we downsample a random class; this is a common practice to inject anomalies by making a class appear as a rare event \citep{Ntroumpogiannis22}. 

\textbf{Evaluation Metric.} The detection performance is assessed using the F1-Score. In our experiments, the noise ratio is considered known, as the data glitch injection is artificial, thus the threshold to compute the F1 is automatically set. 
% Note that we do not suggest a fine-grained threshold method but we show whether glitches can be accurately detected.

\textbf{ML Models.} For tabular datasets, we employ FT-Transformer \citep{gorishniy2021revisiting}, a ResNet model as proposed in \citep{gorishniy2021revisiting}, and MLP. The learning rate, number of epochs, batch size, and validation accuracy for each model can be found in our code repository. Each model was fine-tuned using stochastic gradient descent with no momentum to meet TracIn's assumptions \citep{pruthi2020estimating}. For vision datasets, we employ state-of-the-art foundation models with diverse architectures such as ResNet-20 \citep{he2016deep} pre-trained on CIFAR-10, ConvNeXt \citep{liu2022convnet} pre-trained on ImageNet and Vision Transformer (ViT-16B) \citep{DosovitskiyB0WZ21} pre-trained on JFT-300M. Each model was then trained for a few fine-tuning steps on each contaminated training set, minimizing the cross-entropy loss and achieving a good validation accuracy. For each image dataset, we fine-tune a foundation model. Regarding the ResNet-20 - CIFAR-10 combination, the rationale is to continue the training procedure, to increase the predictive performance further while injecting data glitches into the training set. This simulates the scenario of fresh incoming samples with data quality issues that become part of the training set. 

% We employ foundation models with different architectures such as Resnet-20 \citep{he2016deep} pretrained on CIFAR-10, ConvNext \citep{liu2022convnet} trained on ImageNet and Vision Transformer (ViT-16B) \citep{DosovitskiyB0WZ21} trained on JFT-300M. Each model was fine-tuned for 8 epochs, achieving a good validation accuracy when trained with a ``glitched'' dataset \mycomment{If the accuracy plots can fit in the pages, I think we should in include them or say something for the performance}. Please see in Section \ref{sec:dynamic} the rationale behind the few fine-tuning steps. Finally, to meet the assumptions of TracIn \citep{pruthi2020estimating} each model was fine-tuned using stochastic gradient descent with no momentum.

\textbf{IF.} Our empirical study rely on TracIn  \citep{pruthi2020estimating} implemented in the Captum \footnote{\url{https://captum.ai/api/index.html}} package implemented in PyTorch. To speed up computations, the influence is estimated based only on the weights of the last layer of each model, which according to previous studies does not degrade influence estimation \citep{relatif,pruthi2020estimating,nguyenCBIF}.

%%%%%%%%%%%%%%%%%%%%%%%%%%%%%%%%%%%%%%%%%%%%%%%%
%%%% Experimental Evaluation
%%%%%%%%%%%%%%%%%%%%%%%%%%%%%%%%%%%%%%%%%%%%%%%%
\section{Experimental Evaluation}
\label{sec:experimental-evaluation}

In this section, we report the results of the experimental comparison of the influence-based signals for debugging mislabeled and anomalous samples during model training. Note that each experiment was repeated five times with different random seeds. The evaluation pipeline is depicted in Fig. \ref{fig:eval-pipeline} in the Appendix. Note that influence-based glitch detection does not explicitly solve an anomaly or mislabeled detection problem. It solves a classification problem and the influence signals aim to spot data glitches as the model is trained. 

\begin{figure}[t]
  \centering
     \begin{subfigure}[b]{0.30\linewidth}
         \centering
        \includegraphics[width=\textwidth]{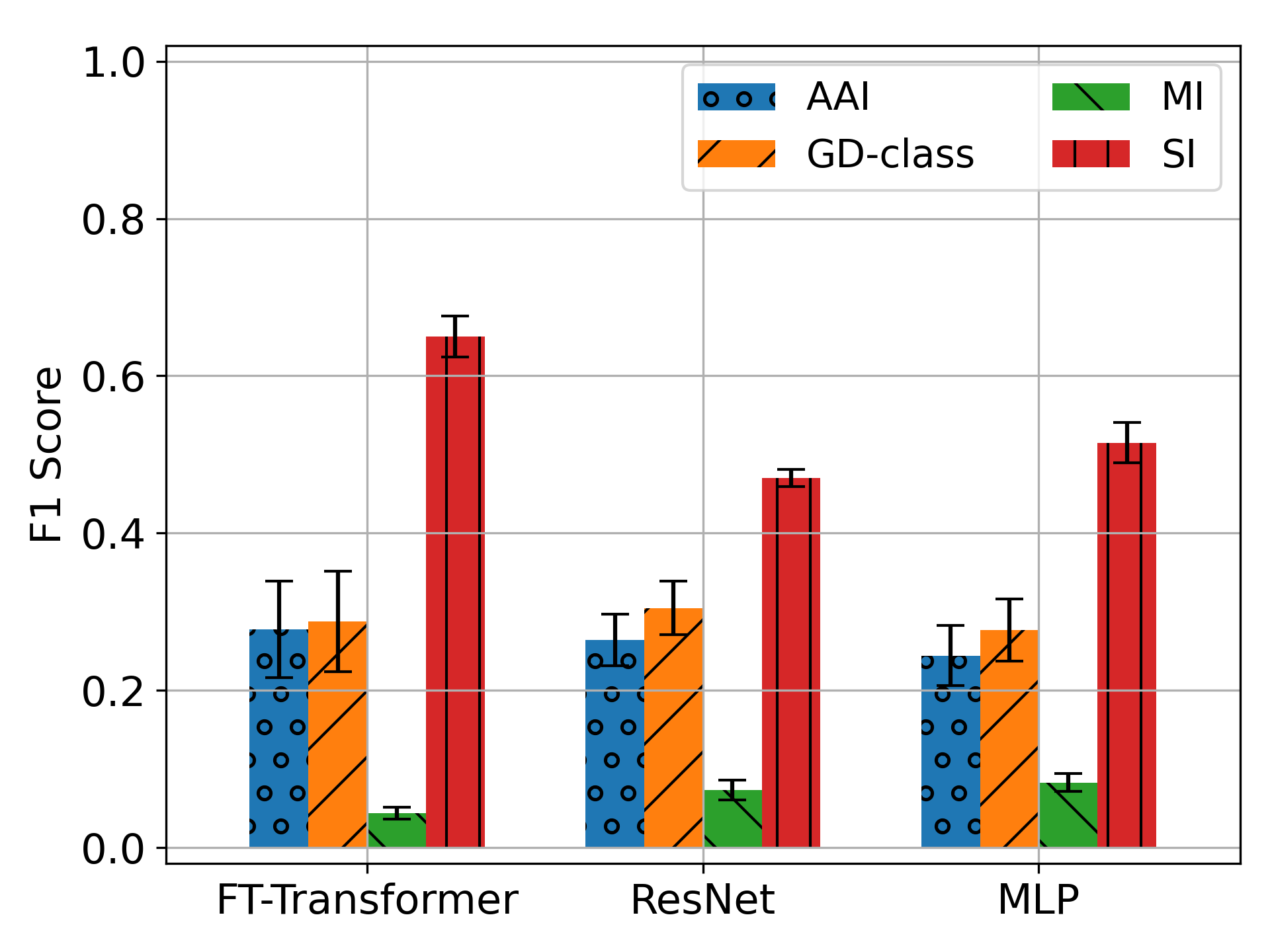}
         \caption{Uniform class noise}
         \label{fig:mu-sigs-tab}
     \end{subfigure}
     \hfill
     \begin{subfigure}[b]{0.30\linewidth}
         \centering
         \includegraphics[width=\textwidth]{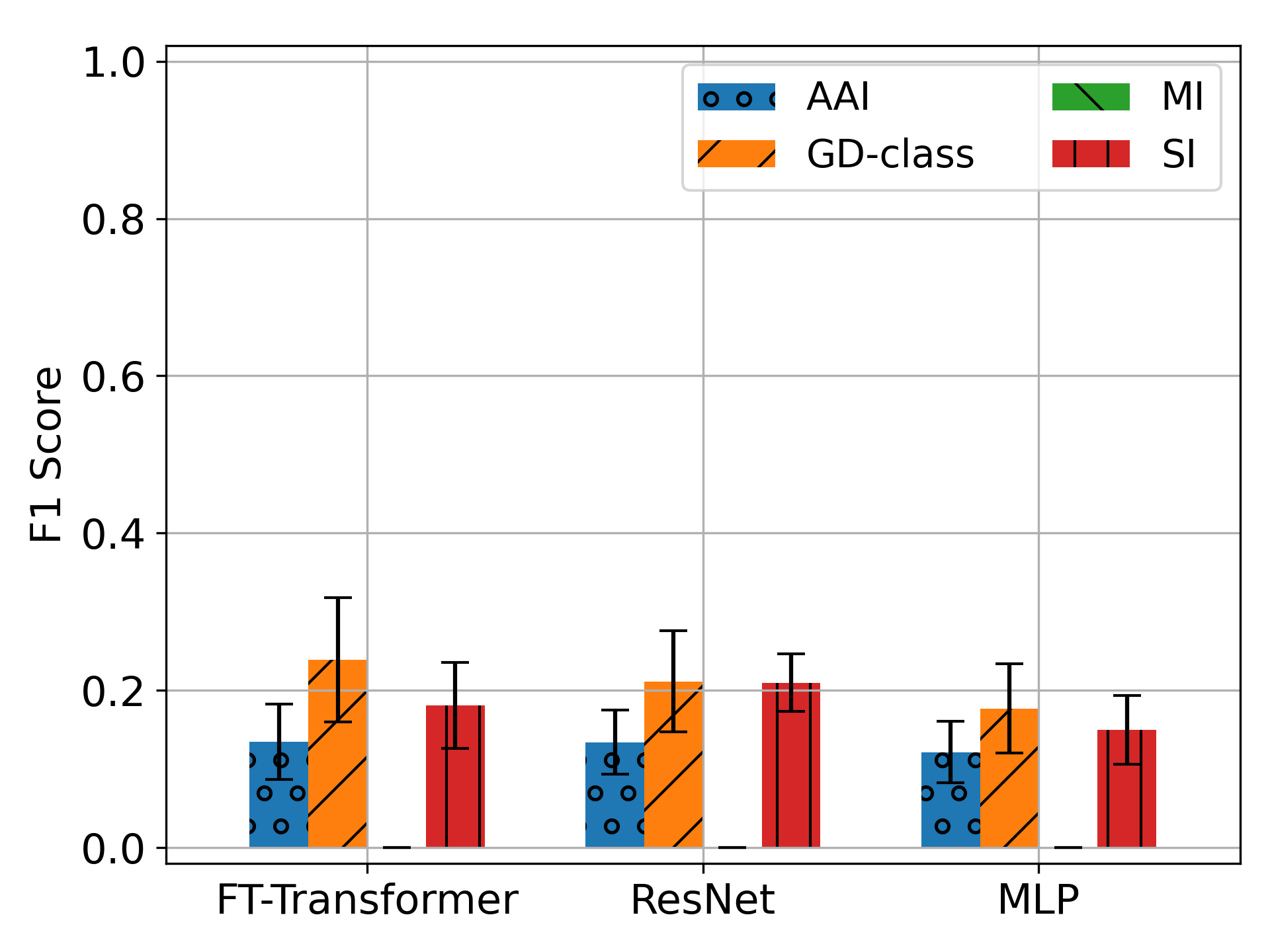}
         \caption{Singular mislabeled class}
         \label{fig:mcb-sigs-tab}
     \end{subfigure}
     \hfill
     \begin{subfigure}[b]{0.30\linewidth}
         \centering
         \includegraphics[width=\textwidth]{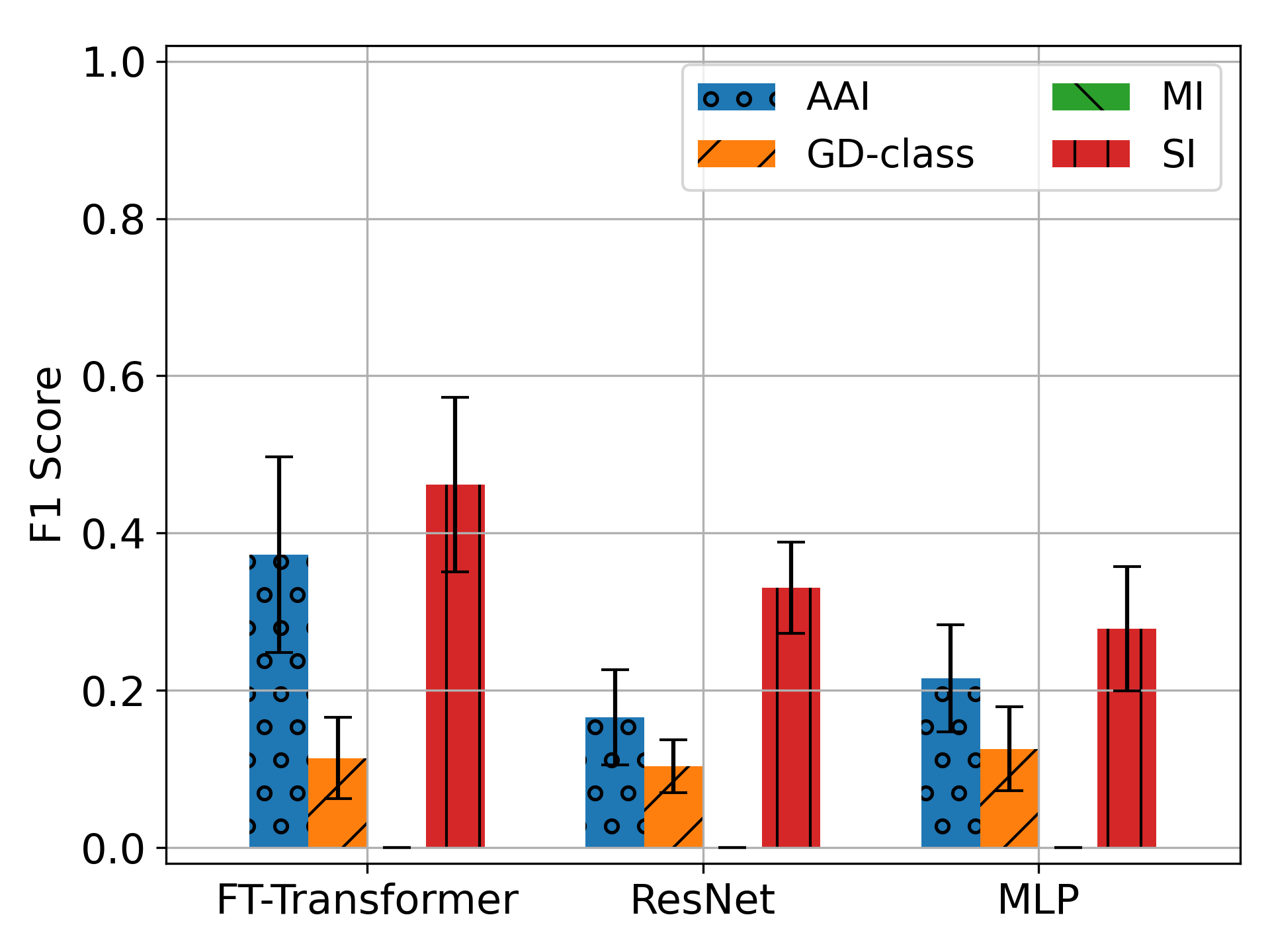}
         \caption{Near-CA}
         \label{fig:anom-sigs-tab}
     \end{subfigure}
     \caption{Detection performance of influence-based signals for mislabeled samples and Near-Clustered Anomalies (Near-CA) on tabular data. All training datasets are contaminated with a 10\% glitch ratio. Each DL model was trained from scratch in all datasets. The average F1-Score is reported.}
        \label{fig:sigs-tab-detection}
\vspace{-.5cm}
\end{figure}

\begin{figure}[ht]
  \centering
     \begin{subfigure}[b]{0.22\linewidth}
         \centering
        \includegraphics[width=\textwidth]{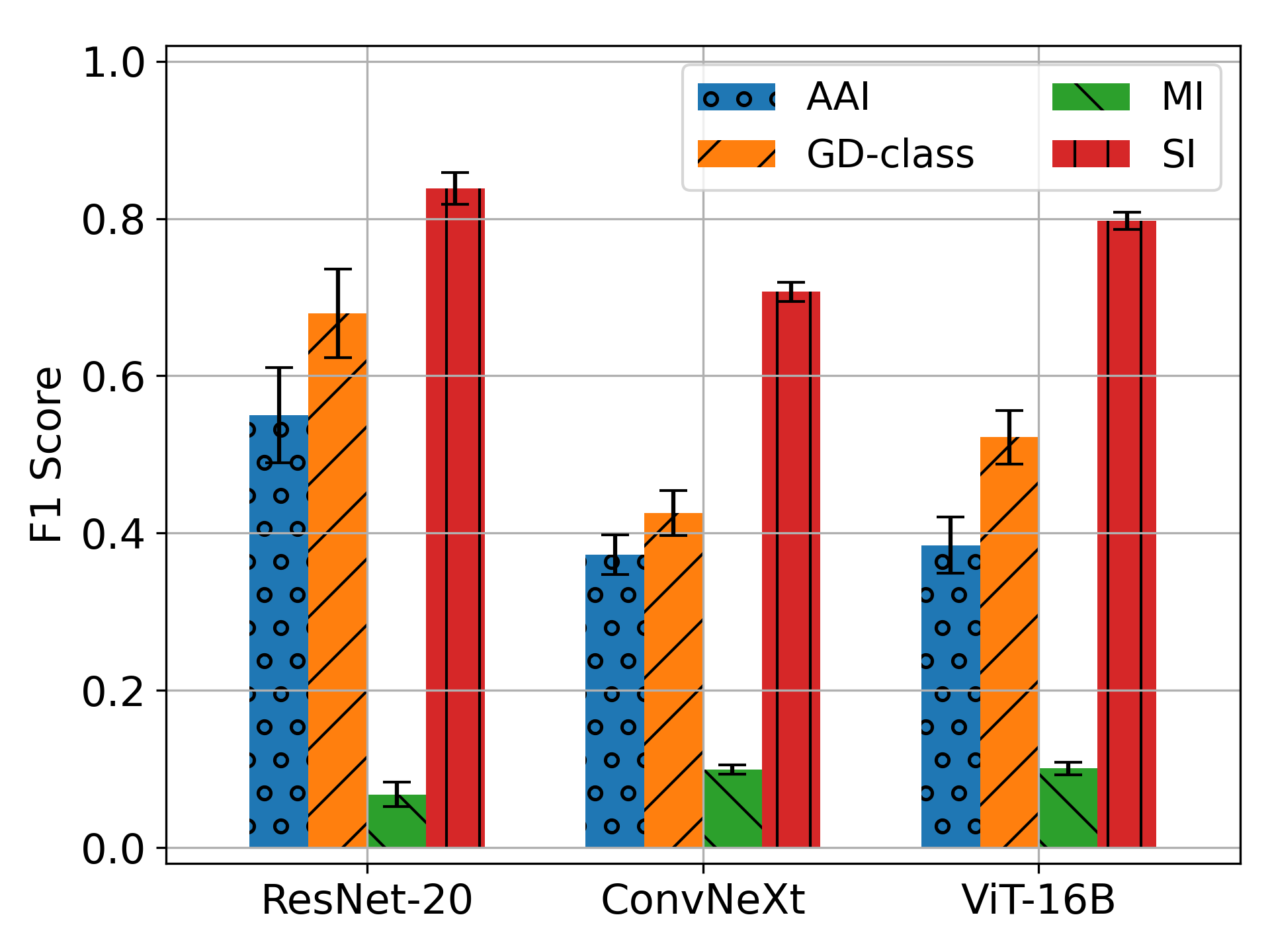}
         \caption{Uniform class noise}
         \label{fig:mu-sigs}
     \end{subfigure}
     \hfill
     \begin{subfigure}[b]{0.22\linewidth}
         \centering
         \includegraphics[width=\textwidth]{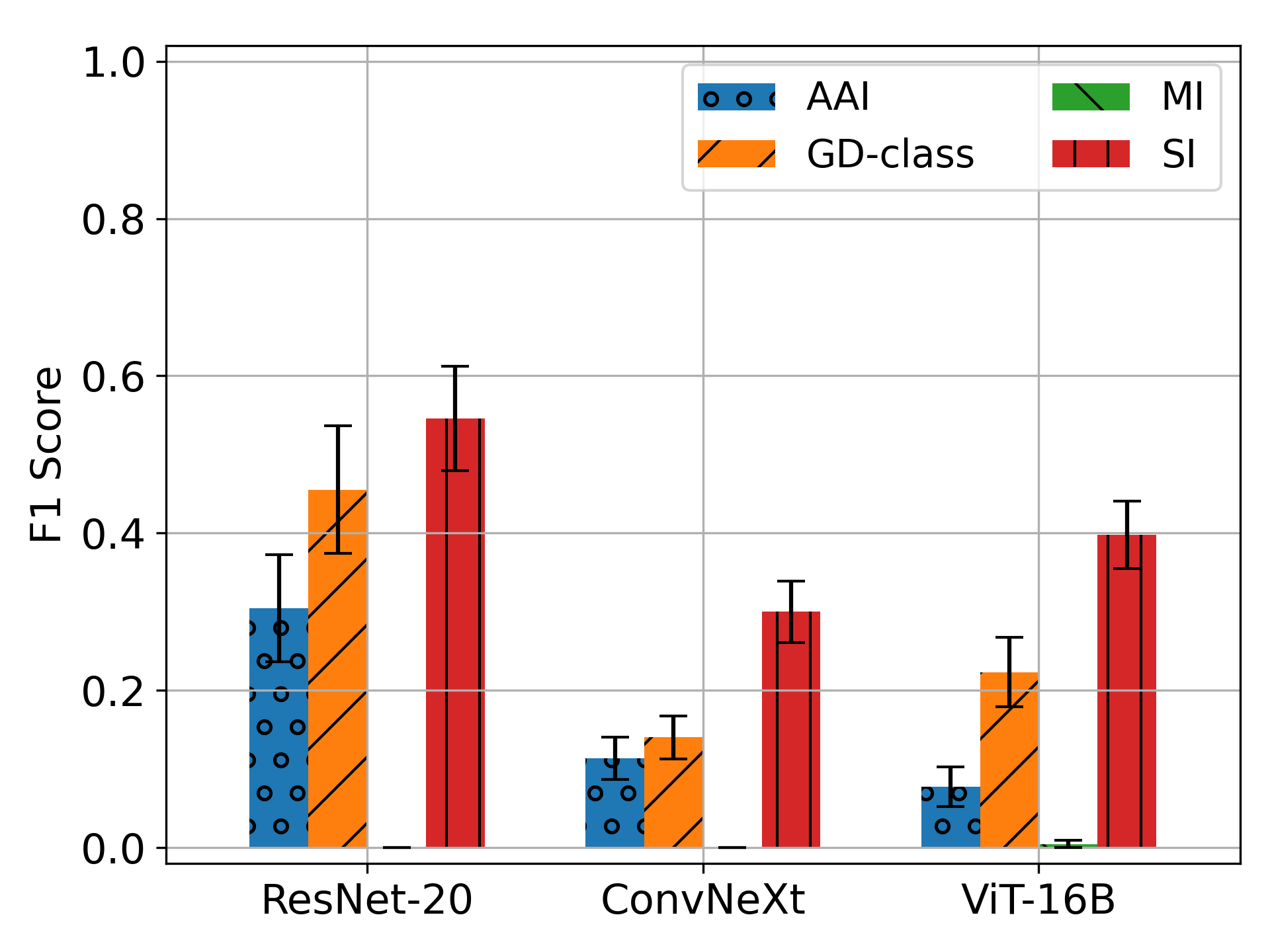}
         \caption{Singular mis. class}
         \label{fig:mcb-sigs}
     \end{subfigure}
     \hfill
     \begin{subfigure}[b]{0.22\linewidth}
         \centering
         \includegraphics[width=\textwidth]{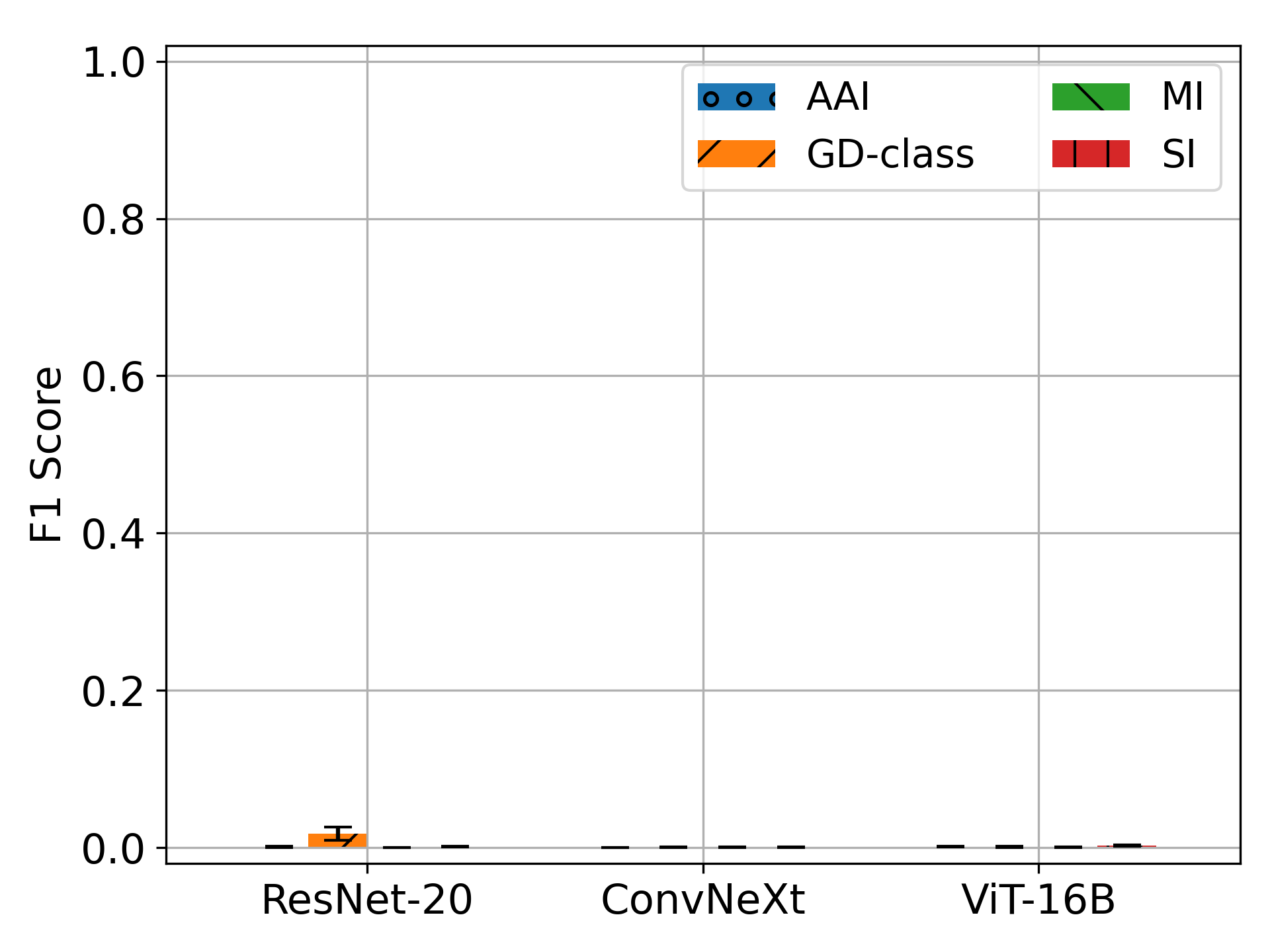}
         \caption{Far-CA}
         \label{fig:anom-sigs}
     \end{subfigure}
     \hfill
     \begin{subfigure}[b]{0.22\linewidth}
         \centering
         \includegraphics[width=\textwidth]{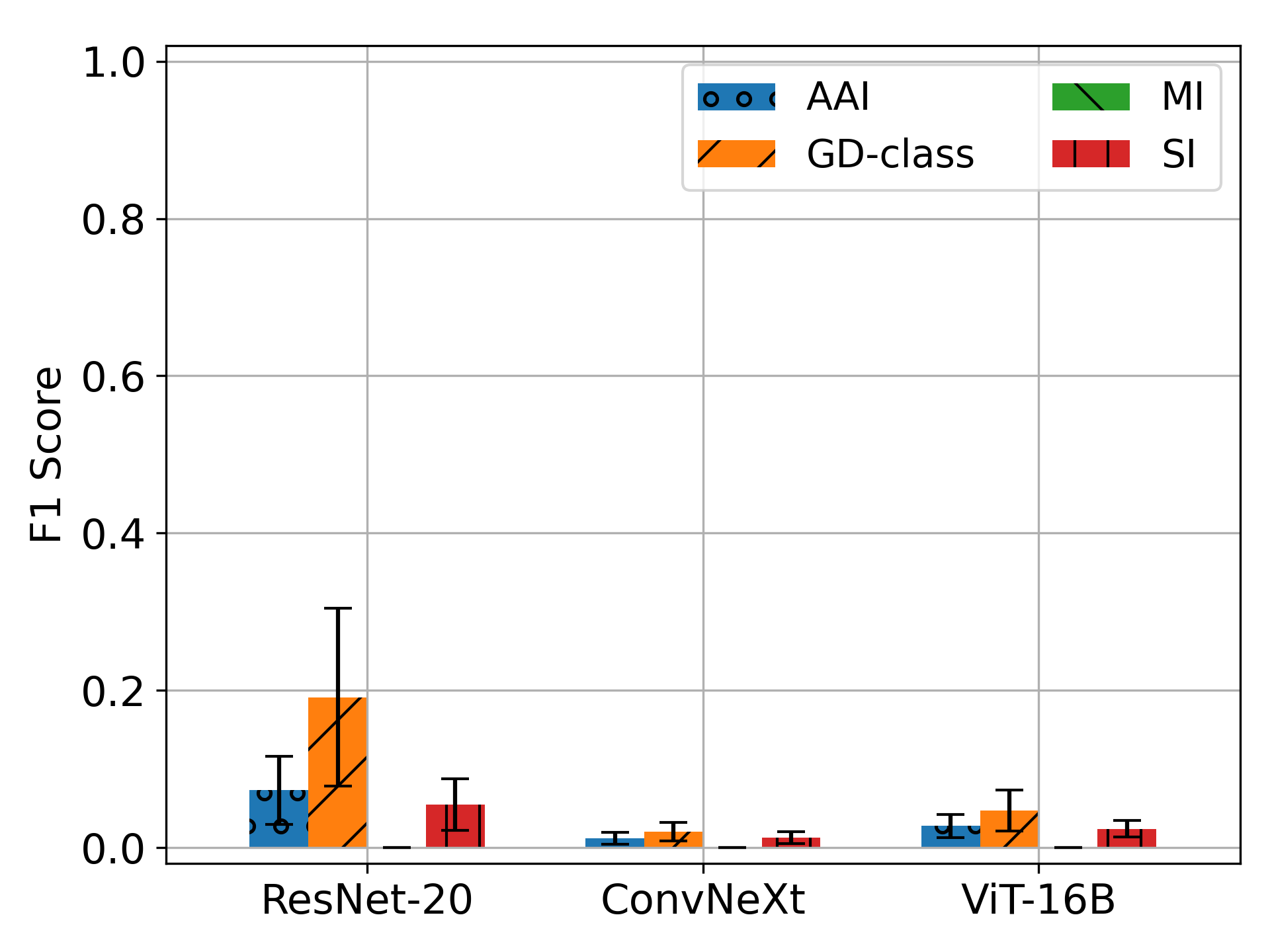}
         \caption{Outliers}
         \label{fig:outlier-sigs}
     \end{subfigure}
     \caption{Detection performance of influence-based signals for mislabeled (mis.) samples, Far-Clustered Anomalies (Far-CA), and outliers on image data. All training datasets are contaminated with a 10\% glitch ratio. Each vision foundation model was fine-tuned in all datasets. The average F1-Score is reported.}
    \label{fig:sigs-image-detection}
\vspace{-.5cm}
\end{figure}

%\subsection{Data Debugging using Influence Signals}
As depicted in Fig. \ref{fig:mu-sigs-tab} and \ref{fig:mu-sigs}, SI is the most effective signal for detecting uniform class noise and generalizes well across different DL models for tabular (trained from scratch) and image data (foundation). Changing the mislabeled ratio seems to have an impact on detection performance as depicted in Fig. \ref{fig:ablation-mu-tab}, \ref{fig:ablation-mu-vis}. For smaller ratios, SI performance decreases, while for higher ratios, it increases. When only one class is mislabeled (singular mislabeled), i.e., fewer mislabeled samples, SI's performance decreases substantially, especially when DL models are trained from scratch. Overall, the influence signals are more effective when detecting class noise on foundation models in image datasets.

Regarding clustered anomalies, Near-CA seem to cause more substantial parameter updates than the clean samples on average, resulting in high SI and AAI. However, as depicted in Figure \ref{fig:ablation-anom-tab} in the Appendix, the SI and AAI signals are able to detect Near-CA samples only in the Epsilon dataset which is a binary classification dataset. The F1-Score drops close to zero for multi-class datasets (Jannis and Forest Cover). The difference in the detection performance for binary and multi-class tabular datasets motivated us to dive deeper into the training dynamics. As shown in Fig. \ref{fig:f1-per-epoch} in the Appendix, SI can detect more accurately the Near-CA samples in early epochs for both datasets in FT-Transformer. This highlights the need for the development of a new class of signals that consider the training dynamics, i.e., selecting which epochs to choose to detect Near-CA samples.  It is worth noting that according to Table \ref{tab:ca-acc}, all the employed models achieve close to zero accuracy on Near-CA samples, i.e., none of the models learn to classify them. This is not the case for the Far-CA samples in the image datasets. As shown in Table \ref{tab:ca-acc}, ResNet-20 as well as the rest of the foundation models can classify Far-CA samples with 100\% accuracy. Interestingly enough, the classification performance difference does not lead to better detection performance from the influence signals. As depicted in \ref{fig:anom-sigs} all existing signals fail to separate Far-CA from clean samples. Finally, the existing signals are also not effective in spotting outliers as illustrated in Fig. \ref{fig:outlier-sigs}. 

\emph{Lessons Learned.} SI is the most effective signal for detecting uniform class noise but its performance drops for small noise ratios or when a singular class is contaminated. MI exhibits the worst performance due to the influence cancellation effect. Class unsigned influence scores as used by GD-class are not enough to outperform SI. Finally, none of the signals can detect Near-CA for multi-class tabular datasets, Far-CA, and outliers for the image datasets. However, when training dynamics are considered, SI can detect Near-CA samples even for multi-class datasets.

\section{Conclusion and Open Challenges}
\label{sec:discussion}
Our analysis highlights that existing signals such as Self-Influence can effectively identify uniform class noise, especially for vision foundation models. However, none of the existing signals can identify clustered anomalies such as Near-CA, Far-CA and outliers that are formed via corruptions. We stress that influence-based glitch detection requires both (i) methodological and (ii) foundational advances. Regarding (i), our experimental findings suggest that new joint influence signals should perform \emph{sign}-wise aggregations to avoid the influence cancellation effect. Moreover, signals should leverage the training dynamics to potentially discover glitches in the early epochs of the training rather than rely on the aggregation of the influence scores (e.g. from TracIn). Regarding (ii), new influence signals are needed to detect different types of anomalies. 

Despite the substantial efforts on the development of more accurate influence estimators, it remains open how one can leverage influence scores to form informative signals that assist analysts in accurately debugging \emph{training} sets and characterizing the \emph{type} of data imperfections especially when mixtures of data glitches may co-exist. Glitch characterization is particularly important as the type of data imperfections is not known beforehand. Finally, data debugging systems should additionally suggest repairs for data glitches such as mislabeled samples. We believe that more research needs to be devoted to the design of \emph{Explainable Data Debugging Frameworks} that are able to detect, characterize, and repair data glitches as we train a target ML model, and hence alleviate the upstream use of dedicated detectors of mislabeled or anomalous samples.

% \impact{Authors must include a Broader Impact Statement, which should provide a concise, tangible portrayal of both the potential positive and negative societal consequences of their work. We refer to the submission guidelines for further details.
% }

% Acknowledgements and Disclosure of Funding should go at the end, before appendices and references

% \acks{All acknowledgements go at the end of the paper before appendices and references.
% Moreover, you are required to declare funding (financial activities supporting the
% submitted work) and competing interests (related financial activities outside the submitted work).
% More information about this disclosure can be found on the DMLR website.}

\vskip 0.2in
\bibliography{literature}
\clearpage

%%%%%%%%%%%%%%%%%%%%%%%%%%%%%%%%5
%% APPENDIX
%%%%%%%%%%%%%%%%%%%%%%%%%%%%%%%%%%

\appendix

\section{Evaluation Pipeline}

\begin{figure}[h]
    \centering
    \includegraphics[width=1.0\textwidth]{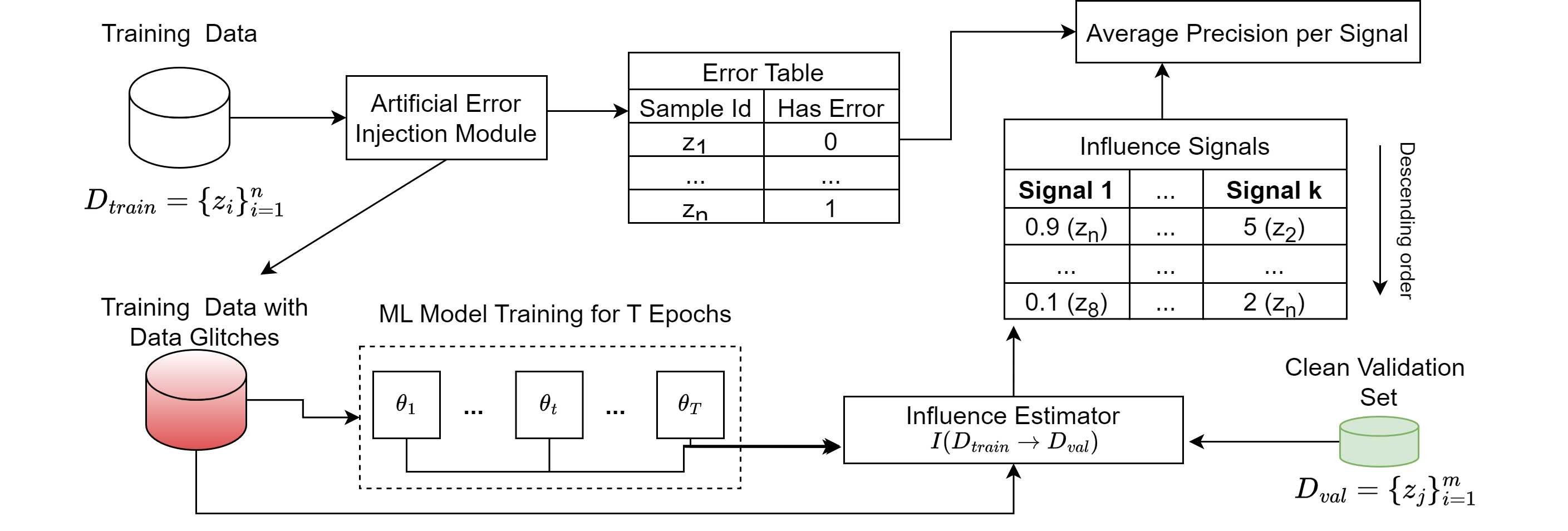}
    \caption{Evaluation Pipeline of Influence-based Signals}
    \label{fig:eval-pipeline}
\end{figure}

The training pipeline is depicted in the Fig. \ref{fig:eval-pipeline}. Given a training dataset, (i) artificial data glitches such as mislabeled, Near-CA, Far-CA, and outliers are injected. This error injection module exports the ``glitched'' dataset and an error table that contains a binary label regarding which sample is glitched. Then (ii) the ML model is trained directly to the ``glitched'' training dataset for $T$ epochs and the the model’s parameters are stored for each epoch. Subsequently, (iii) the IF such as TracIn, reads the saved model checkpoints, and the train-to-validation influence is computed. The influence-based signals perform aggregations on the influence scores, and each training samples receives a score. Finally, their scores are ranked in descending order, as each signal assumes that ``glitched'' samples tend to get higher scores. Finally, the F1-Score is computed using the error table and the scores of each signal. F1-score serves as the detection accuracy and is computed based on the known glitch ratio.

\section{Ablation Study}
\label{sec:app-ablation}

\begin{figure}
  \centering
     \begin{subfigure}[b]{0.49\linewidth}
         \centering
        \includegraphics[width=\textwidth]{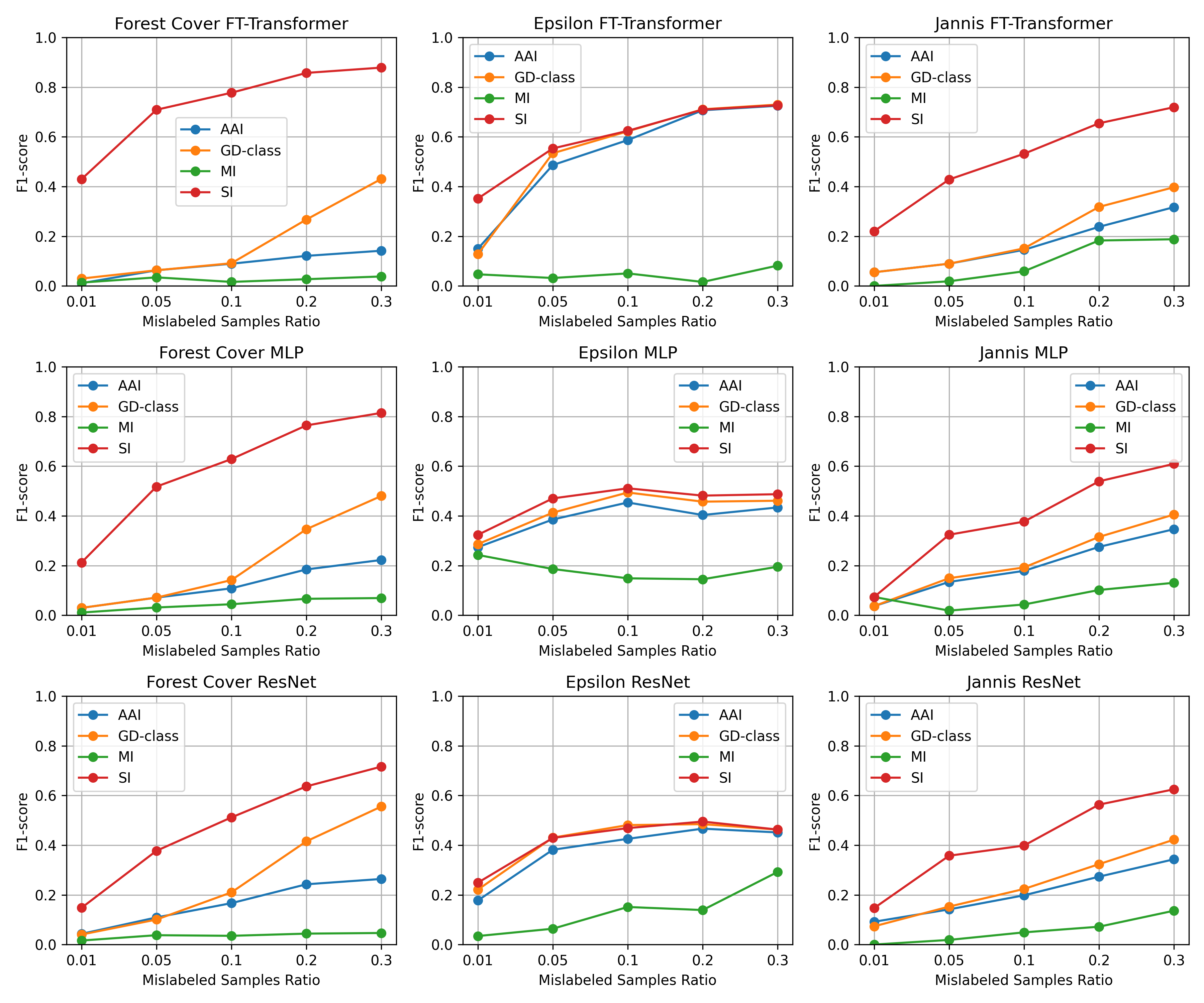}
         \caption{Detection of uniform class noise on tabular data}
         \label{fig:ablation-mu-tab}
     \end{subfigure}
     \hfill
     \begin{subfigure}[b]{0.49\linewidth}
         \centering
         \includegraphics[width=\textwidth]{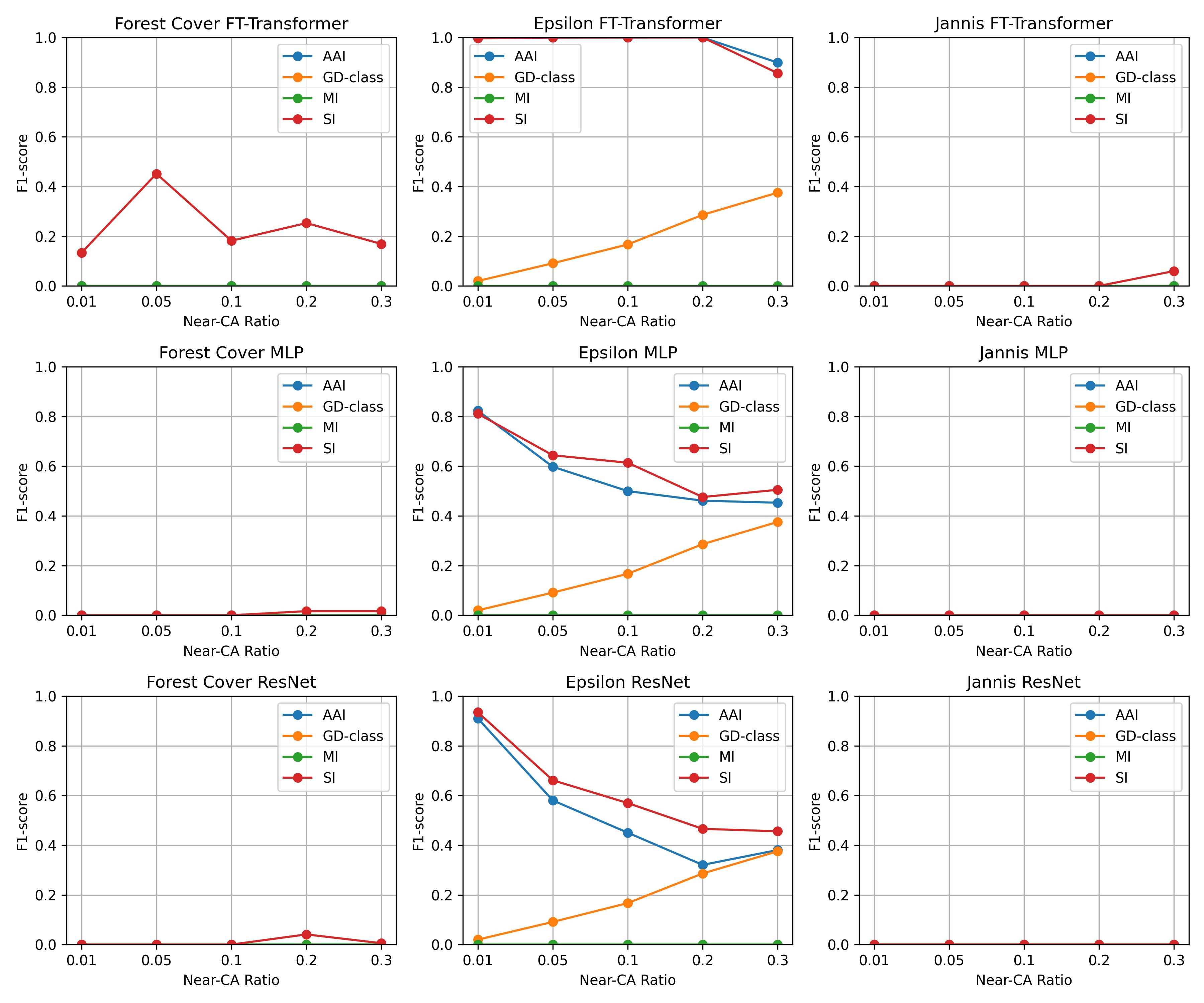}
         \caption{Detection of Near-CA on tabular data}
         \label{fig:ablation-anom-tab}
     \end{subfigure}
     \hfill
     \begin{subfigure}[b]{0.49\linewidth}
         \centering
         \includegraphics[width=\textwidth]{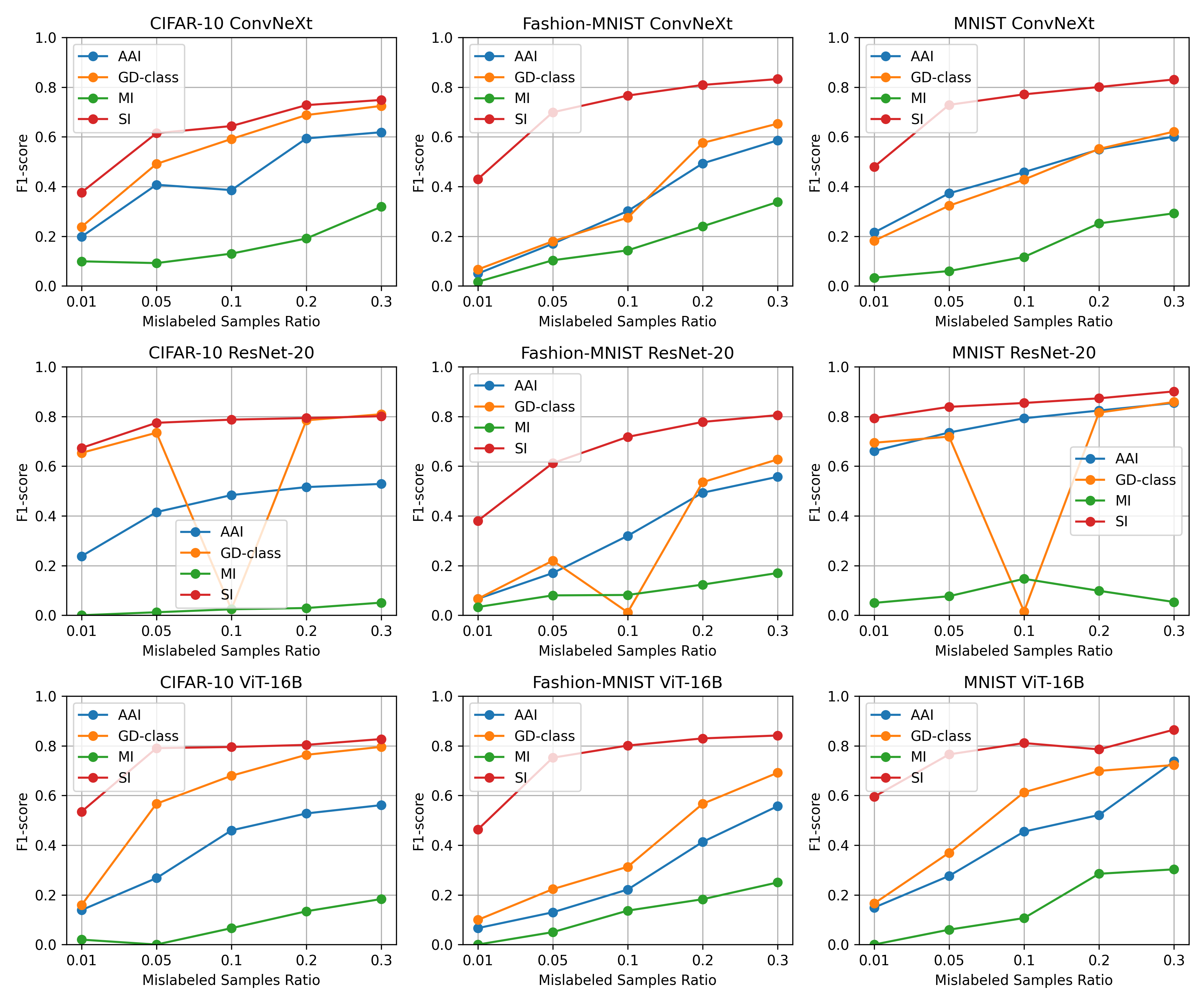}
         \caption{Detection of uniform class noise on image data}
         \label{fig:ablation-mu-vis}
     \end{subfigure}
      \hfill
     \begin{subfigure}[b]{0.49\linewidth}
         \centering
         \includegraphics[width=\textwidth]{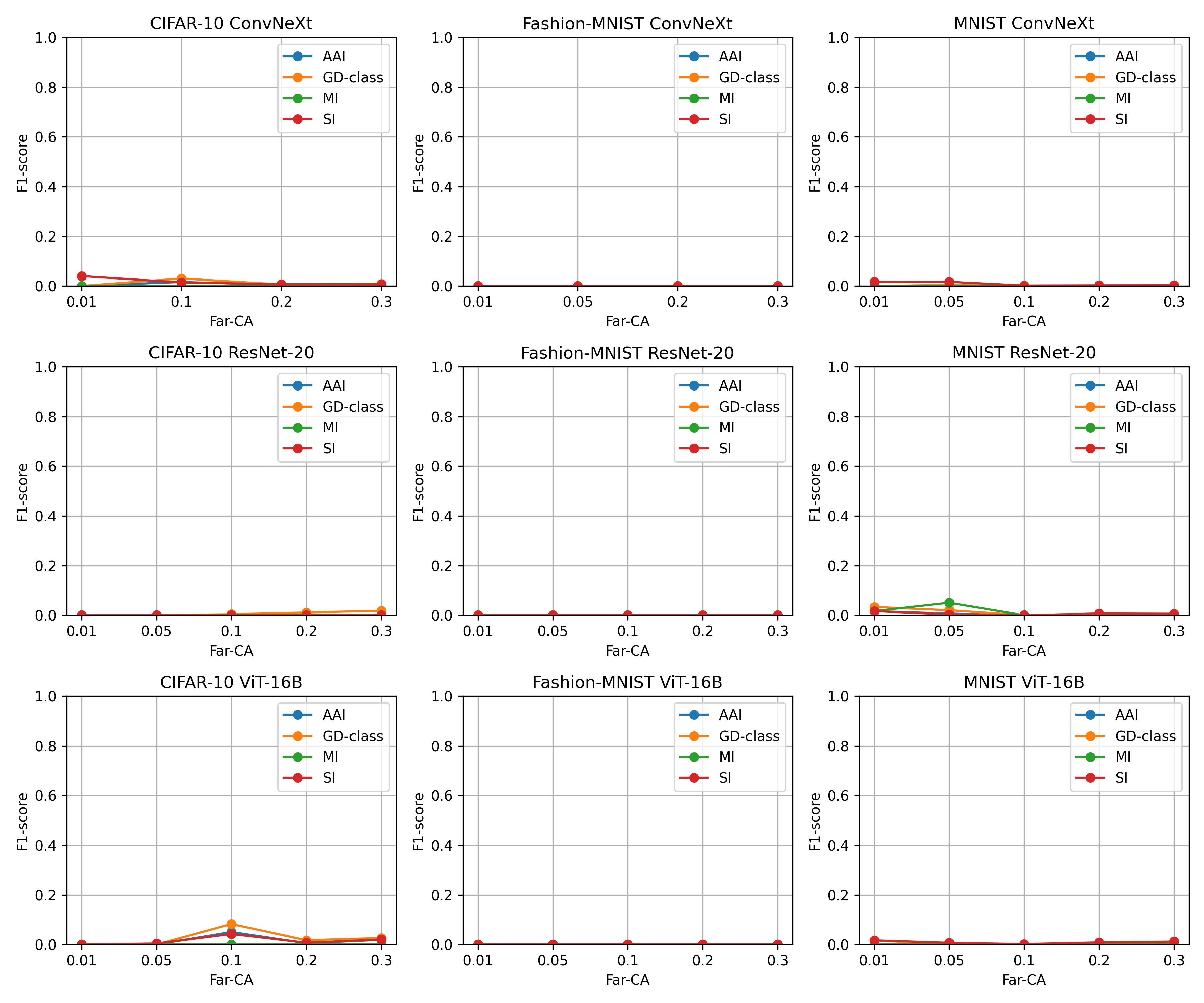}
         \caption{Detection of far-CA on image data}
         \label{fig:ablation-anom-vis}
     \end{subfigure}
     \caption{Data glitch detection performance of influence signals w.r.t. increasing ratio of uniform class noise (both tabular/image data and from-scratch/foundation models), Near-CA for tabular datasets when DL models are trained from scratch, and Far-CA for image datasets when foundation models are fine-tuned. }
        \label{fig:ablation-all}
\end{figure}

In this section, we perform an ablation study focusing on increasing glitch ratio starting from 1\% up to 30\%. As we can see infigures \ref{fig:ablation-mu-tab} and \ref{fig:ablation-mu-vis}, the influence signals have similar behavior when the DL models are trained from scratch on the tabular datasets and when the foundation models are fine-tuned on the image datasets. Specifically, SI's detection performance tends to decrease for smaller mislabeled ratios and increase for higher ratios. Regarding the Near-CA, the signals detect such samples only in the binary classification dataset Epsilon, while they fail to identify Near-CA samples in multi-class settings respectively of the glitch ratio. Finally, for the Far-CA samples,  

\section{Model's accuracy on clustered anomalies}

In this experiment, we focus on FT-Transformer trained from scratch on each tabular dataset (Jannis, Forest Cover, and Epsilon), and the foundation model ResNet-20 fine-tuned on each image dataset (MNIST, Fashion-MNIST, and CIFAR-10). Table \ref{tab:ca-acc} shows the classification accuracy only for Near-Clustered Anomalies (Near-CA) on tabular datasets and Far-Clustered Anomalies (Far-CA) on image datasets. FT-Transformer is not able to classify correctly Near-CA samples in none of the three datasets. On the contrary, ResNet-20 is able to perfectly learn and classify samples from a completely different distribution as one of the existing classes of the corresponding datasets\footnote{recall that we assign a random label from the existing classes to the Far-CA samples.}. Despite the models' accuracy differences in these samples, none of the signals effectively detect the clustered anomalies in all datasets.

\begin{table}[h]
\begin{tabular}{|c|c|c|c|c|}
\hline
\textbf{Dataset} & \textbf{Modality} & \textbf{Model} & \textbf{Glitch Type} & \textbf{Accuracy on Clustered Anomalies} \\ \hline
MNIST            & Image             & ResNet-20      & Near-CA              & 1                 \\ \hline
Fashion-MNIST    & Image             & ResNet-20      & Near-CA              & 1                 \\ \hline
CIFAR-10         & Image             & ResNet-20      & Near-CA              & 1                 \\ \hline
Forest Cover     & Tabular           & FT-Transformer & Far-CA               & 0                 \\ \hline
Jannis           & Tabular           & FT-Transformer & Far-CA               & 0                 \\ \hline
Epsilon          & Tabular           & FT-Transformer & Far-CA               & 0.005             \\ \hline
\end{tabular}
\caption{Accuracy of ResNet-20 and FT-Transformer on tabular and image datasets on predicting Near and Far Clustered Anomalous (CA) samples.}
\label{tab:ca-acc}
\end{table}

\section{Training dynamics of signals for Near-CA samples}

\begin{figure}[h]
  \centering
     \begin{subfigure}[b]{0.4\linewidth}
         \centering
         \includegraphics[width=\textwidth]{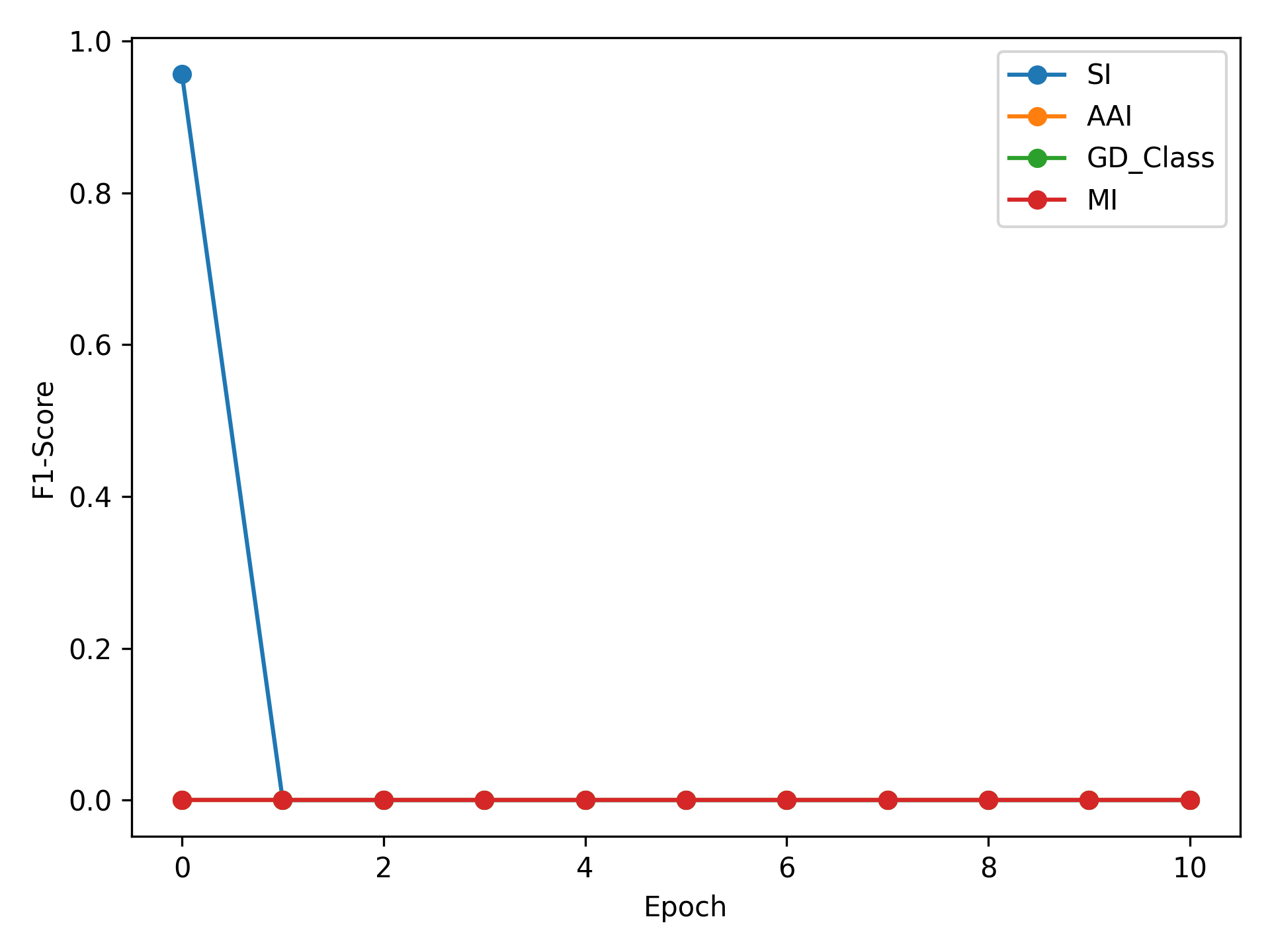}
         \caption{Jannis}
         \label{fig:f1-per-epoch-jannis}
     \end{subfigure}
      \hfill
     \begin{subfigure}[b]{0.4\linewidth}
         \centering
         \includegraphics[width=\textwidth]{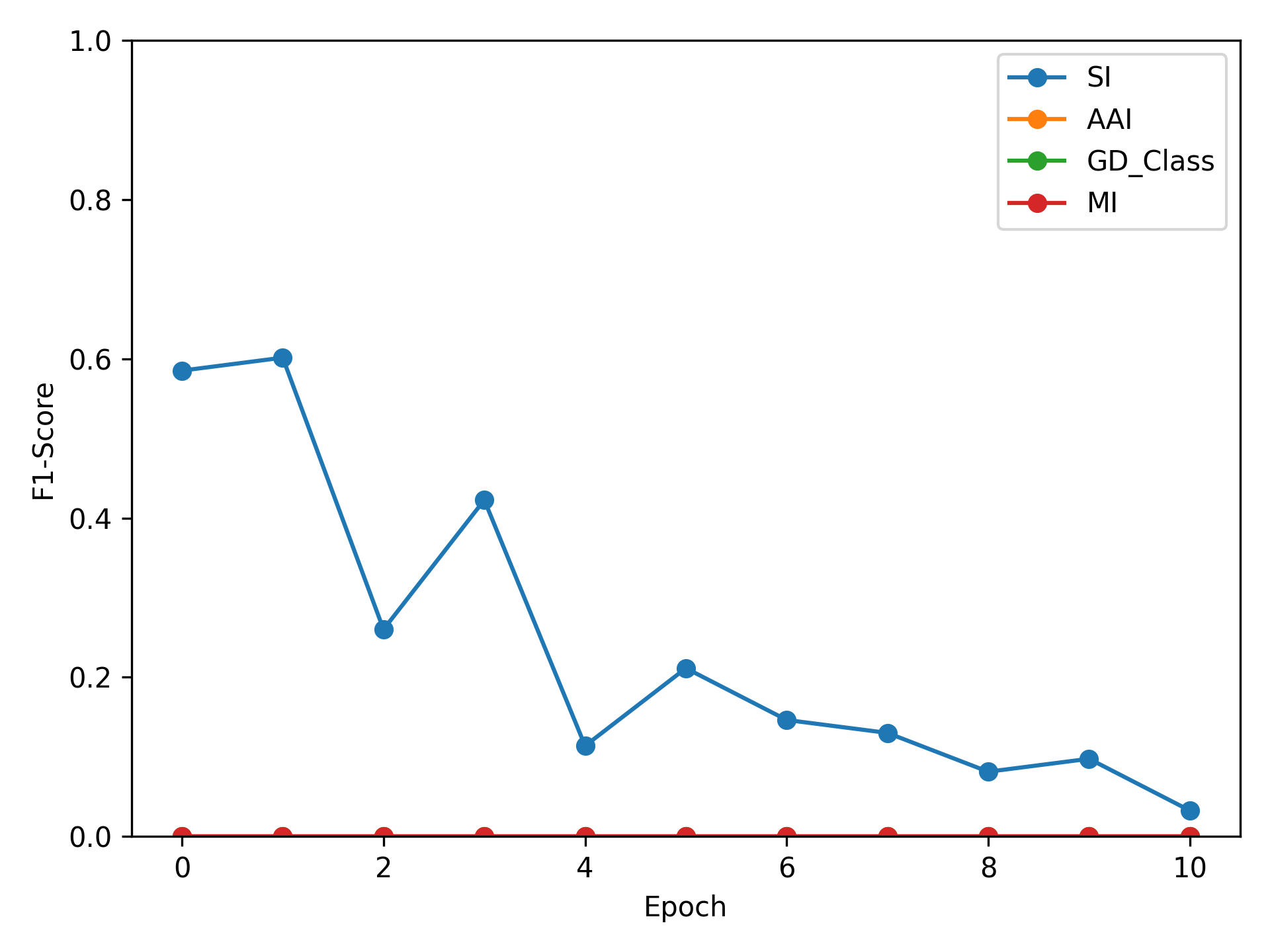}
         \caption{Forest Cover}
         \label{fig:f1-per-epoch-covtype}
     \end{subfigure}
     \caption{Detection performance of influence-based signals for Near-CA reported per epoch when FT-Transformer is trained from scratch on Jannis and Forest Cover}
        \label{fig:f1-per-epoch}
\end{figure}

In this experiment, we report the detection performance for Near-CA samples for Jannis and Forest Cover multi-class tabular datasets of the signals, based on the influence scores of each epoch, i.e., without using the cumulative influence score as defined in TracIn (see \ref{sec:influence-sigs}). We investigate further Jannis and Forest Cover as the detection performance of the signals on these datasets is close to zero (see Fig \ref{fig:ablation-anom-tab}). For the tabular datasets, each model (FT-Transformer, ResNet, and MLP) was trained from scratch for 10 epochs, thus in each epoch, every signal gets a F1-Score. The results are shown in Fig. \ref{fig:f1-per-epoch}. Observe that in the first epoch, Self Influence (SI) spots all Near-CA samples when the FT-Transformer is trained in Jannis. The detection performance is decreased for Forest Cover, starting from 0.6. However, the SI achieves substantially better performance on Forest Cover than the cumulative influence scores aggregation from TracIn.

% Note: in this sample, the section number is hard-coded in. Following
% proper LaTeX conventions, it should properly be coded as a reference:

%In this appendix we prove the following theorem from
%Section~\ref{sec:textree-generalization}:

\end{document}